\documentclass[letterpaper, 10 pt, conference]{ieeeconf}  % Comment this line out if you need a4paper
\IEEEoverridecommandlockouts                              % This command is only needed if you want to use the \thanks command
\usepackage{graphics} % for pdf, bitmapped graphics files
\usepackage{epsfig} % for postscript graphics files
\usepackage{mathptmx} % assumes new font selection scheme installed
\usepackage{times} % assumes new font selection scheme installed
\usepackage{amsmath} % assumes amsmath package installed
\usepackage{amssymb}  % assumes amsmath package installed
\usepackage{hyperref}
\usepackage{url}
\usepackage{multirow}
\usepackage{gensymb}
\usepackage[utf8x]{inputenc}
\usepackage{CJK}
\usepackage{subfigure}
\usepackage{booktabs}
\hypersetup{
    colorlinks=true,
    linkcolor=blue,
    filecolor=magenta,      
    urlcolor=cyan,
}

\title{\LARGE \bf
Discriminative and Semantic Feature Selection for Place Recognition towards Dynamic Environments
}

\author{Yuxin Tian$^{1}$, Jinyu Miao$^{1}$, Xingming Wu$^{*}$, Haosong Yue, Zhong Liu, Weihai Chen
\thanks{$^{1}$Equal contributions.}
\thanks{Authors are with School of Automation Science and Electrical engineering, Beihang University, Beijing, 100191, P. R. China}%

\thanks{*Correspondence Authors: wxmbuaa@163.com}
}

\begin{document}
\begin{CJK*}{GBK}{song} 
\maketitle
%\thispagestyle{empty}
%\pagestyle{empty}

%%%%%%%%%%%%%%%%%%%%%%%%%%%%%%%%%%%%%%%%%%%%%%%%%%%%%%%%%%%%%%%%%%%%%%%%%%%%%%%%
\begin{abstract}

Features play an important role in various visual tasks, especially in visual place recognition applied in perceptual changing environments. In this paper, we address the challenges of place recognition due to dynamics and confusable patterns by proposing a discriminative and semantic feature selection network, dubbed as DSFeat. Supervised by both semantic information and attention mechanism, we can estimate pixel-wise stability of features, indicating the probability of a static and stable region from which features are extracted, and then select features that are insensitive to dynamic interference and distinguishable to be correctly matched. The designed feature selection model is evaluated in place recognition and SLAM system in several public datasets with varying appearances and viewpoints. Experimental results conclude that the effectiveness of the proposed method. It should be noticed that our proposal can be readily pluggable into any feature-based SLAM system.

\end{abstract}

% Write in corresponding tex files for each section
% Do not write here
\section{Introduction}
\label{sect:intro} 
Simultaneous Localization and Mapping (SLAM) \cite{slam} refers the ability of robot that localizes itself and meanwhile, incrementally build a map of environments during exploration. Visual place recognition plays an essential part in (re-)localization and loop closure detection procedure of SLAM system. It can localize the query image by retrieving the most similar images cached in a pre-built map so that it can estimate the current pose of robot, and then correct the accumulated drifts after long-term navigation and mapping.

Autonomous robots aiming at long-term exploration undergo various visual interference, such as changing weathers, repetitive textures, dynamics, etc. Embedding images into appearance-invariant features is helpful to retrieve the same place and distinguish similar looking ones under varying visual conditions, and thus necessary for appearance-based place recognition and SLAM. Traditional hand-crafted features are always divided into two categories, namely, global features \cite{gist,histofcolor,vlad,hog} and local features \cite{sift, surf, orb}. In general, global features are computational efficient in retrieval but sensitive to varying viewpoints and dynamic occlusions. On the contrary, local features are more robust against viewpoint and appearance changes. They detect hundreds of interest points, also called keypoints, and describe the visual information of neighboring area around each keypoint by a vector, named descriptors. Efficient place recognition algorithms \cite{dbow, bocnf, ibuild} usually firstly extract reliable local features from current frame and then embed them into a global vector \cite{vlad, bow} for the balance between efficiency and accuracy. Such a scheme is the theoretical core of Bag-of-Word-based (BoW-based) algorithms \cite{dbow} and popularly used in practical SLAM systems \cite{orbslam, orbslam2, orbslam3}. However, the performances of existing traditional features degrade when suffering from challenging interference in the real environments.

With the incredible success of deep neural networks in various computer vision tasks, many researchers \cite{superpoint, d2, r2d2,netvlad} have arisen interests in applying deep Convolutional Neural Networks (CNNs) to achieve marvelous improvements. These learned counterparts perform better on image retrieval or matching tasks \cite{hpatches}, but still have difficulties in accurate localization in complex environments. It is due to the neglect of consideration about the stability of regions where features are detected. A stable region means a static, distinct image patch. In the existing feature algorithms, all the regions in images play equal importance for the feature extraction and they simply extract features based on the properties of pixel intensity, which is unreasonable. For place recognition, frequently changed items, repetitive textures, and dynamics should not be concerned. Thus, feature selection is a meaningful task to improve the performance.

As a simple solution to select features in static regions, semantic segmentation models \cite{unet, segnet, deeplab} can be applied to obtain pixel-wise semantic labels and then select features with manually designated static property, e.g. walls and buildings. However, the property can be hardly defined \cite{rapnet}. For instance, The moving cars anc parking cars shoule be distinguished, which are frequently occrred in place recognition dataset but can not judged by semantic labels. Motivated by the success of attention mechanism \cite{attention} in computer vision tasks, many proposals about feature selection \cite{crn, delf, lln, rapnet} are proposed and they estimate the importance of regions to generate more robust descriptors or detect more reliable keypoints. By using attention models, regions play different interesting in different tasks and the invariance against changing appearances is improved. However, there are apparent disadvantages of state-of-the-arts methods that they only work when incorperating with specific features, limiting the generalization of feature selection mechanisms.

In this paper, we propose a novel fully convolutional network (FCN), named DSFeat, to estimate a pixel-wise stability of regions, which can provide a reliable guidance to select stable features regardless the feature algorithms. We use both matching and semantic supervision to train our model for better convergence. We provide a compared results of selected features are shown in Fig.~\ref{fig:feature}. It can be clearly seen that all the car in images are estimated as dynamic objects based on semantic labels, which sometimes discard useful visual information in a scene with plenty of parking cars. As a comparison, our DSFeat can effectively distinguish running cars from parking cars and provide more stable features to perform place recognition. Detailed experiments have proven that our propose model can select robust features located in stable regions effectively. The main contributions of this paper can be summarized as follows:

\begin{enumerate}
    \item A FCN model, DSFeat, is proposed to estimate a pixel-wise activation map, which indicates the stability of regions where features are detected;
    \item A hybrid supervision is introduced to train the model that combined strong supervised regression and self-supervised optimization, i.e. manually designated semantic property and metric learning methods;
    \item Selected features are experimentally concluded to be more robust in BoW-based place recognition and overall SLAM system. 
\end{enumerate}

The remaining of this paper is organized as follows. Section~\ref{method} describes the methodology in detail. And comprehensive experiments are shown in Section~\ref{experiments}. Finally, conclusions and future works are discussed in Section~\ref{Conclusion}.

\begin{figure}[!t]
\centering{\includegraphics[width=1\linewidth]{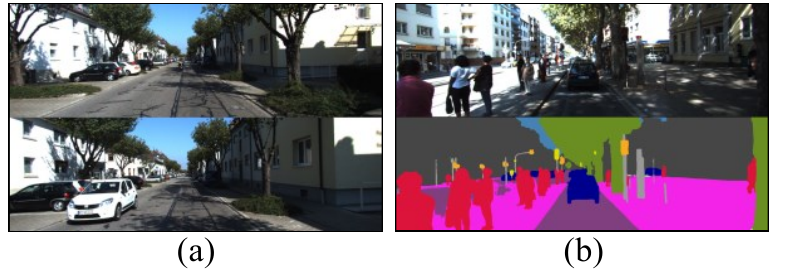}}
\vspace{-0.6cm}
\caption{Examples in the prepared training dataset. (a) shows a pair of matched images, called matching information. And (b) shows a semantic segmentation example, called semantic information.}
\label{fig:datapre}
\end{figure}
\section{Methodology}
\label{method}
In this section, the proposed approach is described in detail. Firstly, we introduce the method of dataset preparation in Section~\ref{dataset preparation}. And in Section~\ref{network structure}, we describe the structure of our model, as illustrated in Fig.~\ref{fig:net}. Finally, the loss function used in our network is introduced in Section~\ref{loss function}.

\subsection{Dataset Preparation}
\label{dataset preparation}
In our proposed method, we apply a hybrid supervision to train our proposed feature selection network, which includes strong supervised regression from semantic segmentation tasks and self-supervised learning from triplets optimization tasks. The construction of triplets need matching relationship between images. Therefore, it is necessary to prepare the semantic information and matching information of the public dataset. 

Matching information in our work means the triplets composited of a query image ($I_q$), a positively matching image ($I_p$), and a negatively matching image ($I_n$). Matched images are caught from the same place, that is, a loop closure in the SLAM system. To obtain accurate matches, we refer to the method \cite{yue} to detect loops. We extract SuperPoint \cite{superpoint} features from the images and apply BoW-based loop closure algorithm to coarsely detect loop candidates. Then topological graph is constructed to verify the candidates. For better accuracy, we manually check the obtained matches to make sure all the matches are definitely from the same scene. These matched loops describe the same place under different visual conditions, as shown in Fig.~\ref{fig:datapre}(a), which includes pedestrians, varying illuminations, and moving cars, etc. In addition, we also need mismatches to construct triplets. For each query image, an image taken in different places, or not passing the graph verification, is considered as the mismatch, i.e. negatively matching image $I_n$. 

For semantic information, we simply apply HRNet \cite{hrnet}, a panoramic segmentation network with high performance, to obtain pixel-level semantic labels of images, as shown in Fig.\ref{fig:datapre}(b). We do not consider any improvement about segmentation and it is beyond this work. Then, we manually designate some static categories, like walls and buildings, and the others are defined as dynamics, following the reference in \cite{calc2}. The label of static items is set to 1 and dynamics is 0 to fit the strategy that higher activation means higher probability of semantic stability. After that, we obtain a binary semantic stability map $S \in \mathcal{R}^{H \times W}$, where H and W is the height and width of original frame.

\subsection{Network Structure}
\label{network structure}
Our approach obtain an one-dimensional activation map with value between 0 and 1, and its resolution is the same as the original image. It is theoretically similar to panoramic segmentation. Thus, we decide to use panoramic segmentation model as our basic backbone. Additionally, the trade-off between computational efficiency and accuracy is necessary for real-time SLAM, so we experimentally compare the comprehensive performance of some famous panoramic segmentation network \cite{segnet, unet, deeplab}, as shown later in Section~\ref{model training}, and finally choose U-Net \cite{unet} as our backbone.

\begin{figure}[!t]
\centering{\includegraphics[width=1\linewidth]{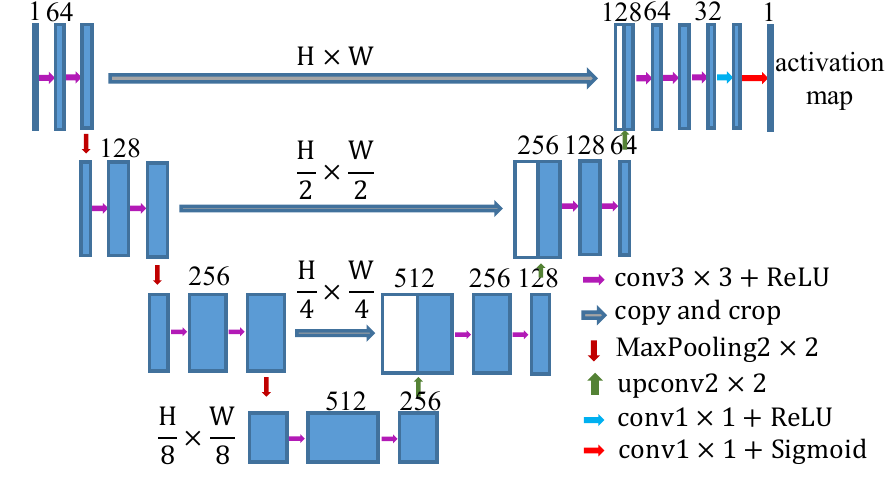}}
\vspace{-0.6cm}
\caption{Improved net structure with U-net as backbone.
\label{fig:net}}
\end{figure}

For achieving better computational efficiency, we improve the vanilla U-net network to adapt to our method. Firstly, the number of \emph{down-sampling} is reduced to improve the speed of network proceeding. Secondly, we increase the depth of the network to improve accuracy. Finally, we use a convolution layer and \emph{Sigmoid} function as the header to get the one-dimension activation map $A \in \mathcal{R}^{H \times W}$. The network structure is shown in Fig.~\ref{fig:net}

\subsection{Loss Function}
\label{loss function}
To train our proposed feature selection model, we use a hybrid loss function (see Section~\ref{hybrid loss}) to optimize the parameters, including semantic loss (Section~\ref{semantic loss}) based on semantic stability and matching loss (Section~\ref{matching loss}) based on matching information of triplets. Additionally, we design two different distance measurement between images based on various characters of dense features and sparse features (Section~\ref{dense distance} and \ref{sparse distance}). Detailed descriptions are listed in following paragraphs.
\subsubsection{Semantic loss}
\label{semantic loss}
The main usage of proposed network is to select stable feature. Existing approaches \cite{lln, rapnet} only use weakly supervised matching information provided by pairs or triplets, and seem to be hard to converge. Therefore, a stronger supervision in the early warm-up stage should be considered. We use semantic segmentation information to achieve such goal. As described in Section~\ref{dataset preparation}, the binary semantic stability labels can be seen as a rough value of activation. Thus, the semantic loss is defined as an initial supervision and we use the binary cross entropy (BCE) function to directly let the activation to regress the semantic stability labels. The semantic loss can be calculated as follows:
\begin{equation}
    \label{eq:1}
    L_{sem}=\frac{1}{wh}\sum_{i=1}^{w}\sum_{j=1}^{h}{BCE(A_{i,j}, S_{i,j})}
\end{equation}
where h and w is the height and width of original input image. $A_{i,j}$ and $S_{i,j}$ indicate the value on (i,j) of $A$ and $S$, respectively.

\subsubsection{Matching loss}
\label{matching loss}
On the other hand, we introduce triplet ranking loss to measure the matching loss. Triplet loss indicates that the dis-similarities between query image and positives should be smaller than those between query and negatives. It can refine the activation map to obtain a more reliable stability and automatically learn to emphasize the stable and distinct regions.
\begin{equation}
    \label{eq:2}
    L_{mat}=max\{d\left(I_{q},I_{p}\right)-d\left(I_{q},I_{d}\right)+m, 0\} 
\end{equation}
where m is a pre-defined parameter and d() is our proposed distance measurement to better indicate the correctness of feature matching after selecting based on activation. As selection procedure is a non-differetiable process, we regard activation as a weight to optimize the network. In order to propose a method which can be readily pluggable into various features, we design different image distance measurement for dense feature and sparse feature, respectively.

\subsubsection{Distance measurement for dense features}
\label{dense distance}
For dense local feature \cite{superpoint, d2, r2d2, sekd}, since feature algorithms provide dense feature map before selection, we can apply an embedding process similar to \cite{lln,rapnet}. We use estimated activation to calculate the weighted mean of dense feature descriptors as a global descriptor, and then measure the similarity between two images based on such descriptors. 
\begin{equation}
    \label{eq:3}
    d\left(I_1,I_2\right)=\frac{1}{wh}||\sum_{i=1}^{w}\sum_{j=1}^{h}{A^{1}_{i,j}d^{1}_{i,j}}-\sum_{i=1}^{w}\sum_{j=1}^{h}{A^{2}_{i,j}d^{2}_{i,j}}||_{2}
\end{equation}
where $d^{a}_{i,j}$ is the descriptor on (i,j) of image $I_{a}$ and $A^{a}_{i,j}$ denotes the activation value on (i,j) of image $I_a$.

\subsubsection{Distance measurement for sparse features}
\label{sparse distance}
For sparse local feature \cite{orb, sift, surf}, its feature have been already selected in a predetermined range and it will affect the convergence of model training since too much information is discard during selection. Thus, we use the activation value as a weight to calculate the weighted mean of matching re-projection error as a stronger supervised distance:
\begin{equation}
    \label{eq:4}
    d\left(I_1,I_2\right)=A\cdot{RE}(I_1, I_2)
\end{equation}
where $RE(I_1, I_2)$ is the re-projection error between image $I_1$ and $I_2$, It can be measured as follows:
\begin{align}
RE(I_1, I_2) = \frac{1}{2n} \sum_{k=1}^{n}(P(F, {u}^1_{i} &, {u}^2_{i})+P(F^{-1}, {u}^2_{i}, {u}^1_{i})) \label{eq:5}\\
P(T, u^a, u^b) =& \frac{(T\times{u}^a)\cdot{u}^b}{||T\times{u}^{a}||} \label{eq:6}
\end{align}
where F is the fundmental matrix from $I_1$ to $I_2$ calculated by random sample consensus (RANSAC) algorithm. ${u}^a_{i}$ is the normalized coordinate of the matching feature point in $I_a$. The function $P$ refers to as the distance between projected epipolar line $T\times{u}_{i}$ and point $u^b$

Fundmental matrix represents the epipolar consistency constraint between images from two viewpoints. Corresponding feature points in the positively matching image pair should meet the constraints of the fundamental matrix, so the re-projection error will be smaller, and verse-visa. Therefore, we can use the average re-projection error as the image distance measurement of sparse feature points for better convergence of the model.

\subsubsection{Hybrid loss}
\label{hybrid loss}
In the end, the loss function we designed includes strongly supervised semantic loss and self-supervised matching loss. Semantic loss offers a rough but easy initial optimization while matching loss provides accurate adaption and refinement. The weighted combination of the two loss is used as the final hybrid loss.
\begin{equation}
    \label{eq:7}
    L=(1-\alpha) L_{sem} +\alpha L_{mat} 
\end{equation}
where $\alpha$ is a hyper-parameter that gradually increases as the training progresses, indicating that matching loss gradually leads the guidance. The hybrid loss function obtains high accuracy results while ensuring good convergence
\section{Experiments}
\label{experiments}
We provide experimental analysis in this section. Firstly, we introduce the details of model training. Then we apply our approach in place recognition tasks to verify the robustness of selected features. Finally, we incorporate our method with the practical slam system to verify the effectiveness of our proposed feature selection mechanism.

\subsection{Model Setups}
\label{model training}
To select a backbone that balances speed and accuracy, we compare the performance of SegNet\cite{segnet}, U-net\cite{unet}, and DeepLab\cite{deeplab} in multiple panoramic segmentation datasets. The result is shown in Table~\ref{table:segment}. For higher processing speed, we finally choose U-net as the backbone for our method.

\begin{table}[t]
\caption{Segmentation backbone comparision}
\label{table:segment}
{
{\begin{tabular}{@{\extracolsep{\fill}}ccccc}
\toprule
\multirow{2}{*}{IOU/FPS} & \multicolumn{4}{c}{Evaluation Dataset} \\ 
& CamCid & CityScapes & SYTNTHIA & Mapillary \\ \midrule
SegNet & 46.4/62 & 70.6/21 & 62.1/44 & 45.4/55 \\
U-net & 44.7/\textbf{85} & 68.4/\textbf{33} & 60.1/\textbf{60} & 44.7/\textbf{75} \\
DeepLab v3 & \textbf{49.5}/10 & \textbf{73.1}/2.6 & \textbf{66.1}/5.1 & \textbf{50.1}/8.3 \\
\bottomrule
\end{tabular}}{}
}
\vspace{-0.2cm}
\end{table}

In the training process, we set the maximal training epoches as 50 to let the network be fully trained. The initial learning rate is set to 1e-3, which will multiply 0.1 in the epoch 20, 30 and 40. In addition, the hyper-parameter $\alpha$ in loss function is initially set as 1 and will multiply 0.9 in every epoch to ensure that semantic loss mainly affects the initial supervision of the network and then matching loss gradually leads the optimization.

\subsection{Evaluation Criteria}
\label{dataset}
We evaluate our method on typical outdoor datasets since dynamic occlusions are frequently occurred in such datasets. CityCentre \cite{citycentre} contains 2474 images and it is caught during 2.0 km exploration in a campus with many similar buildings and shrubs. KITTI odometry \cite{kitti} dataset is one of the most famous binocular benchmark and it contains 22 sequences. 11 sequences (00-10) of them have ground-truth trajectories which can be used for visual odometry or SLAM evaluation.

For place recognition, We use CityCentre and 12 sequences from KITTI to evaluate our methods. Generally used Precision and Recall metrics are measured. Precision refers to as the ratio between correctly detected loops and all the detection while Recall is the ratio between correct detections and all the loop events existing in the current scene. We use ground-truth loops from original publication for CityCentre and manually annotated detections in \cite{dldb} are used as a reference on KITTI dataset.

For SLAM evaluation, we use all of the 11 sequences (00-10) with ground-truth trajectories. CityCentre is not used as we cannot find the public ground-truth poses. We calculate the average error of rotation and the standard deviation of offset between estimation and ground-truth for fair and quantitative comparison.

\subsection{Place Recognition}
\label{matching results}

\begin{figure}[!t]
    \subfigbottomskip=2pt 
  	\subfigcapskip=-5pt
    \subfigure[KITTI-00 +ORB]{
	    \includegraphics[width=0.48\linewidth]{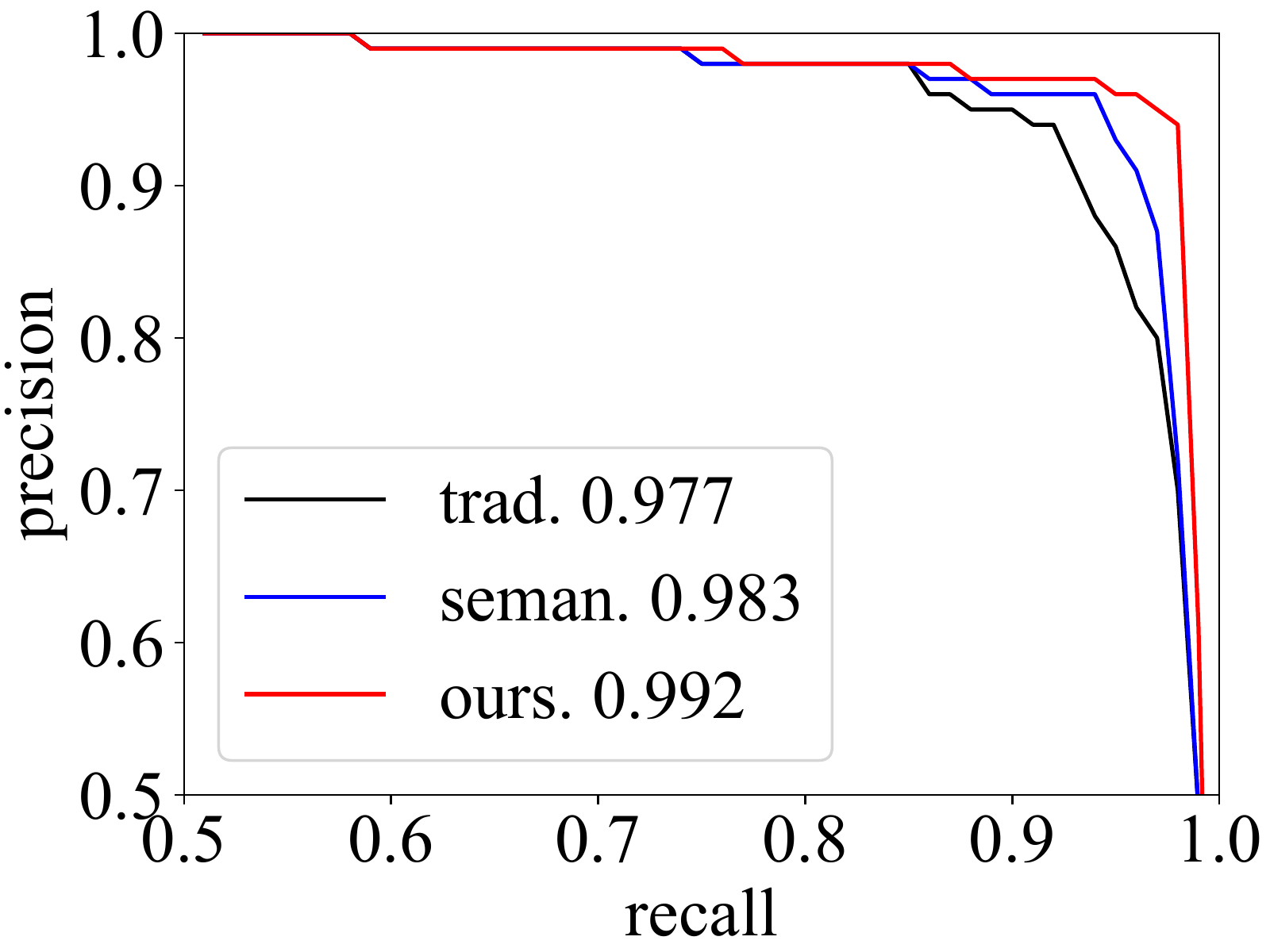}}
	    \subfigure[KITTI-00 +SuperPoint]{
	    \includegraphics[width=0.48\linewidth]{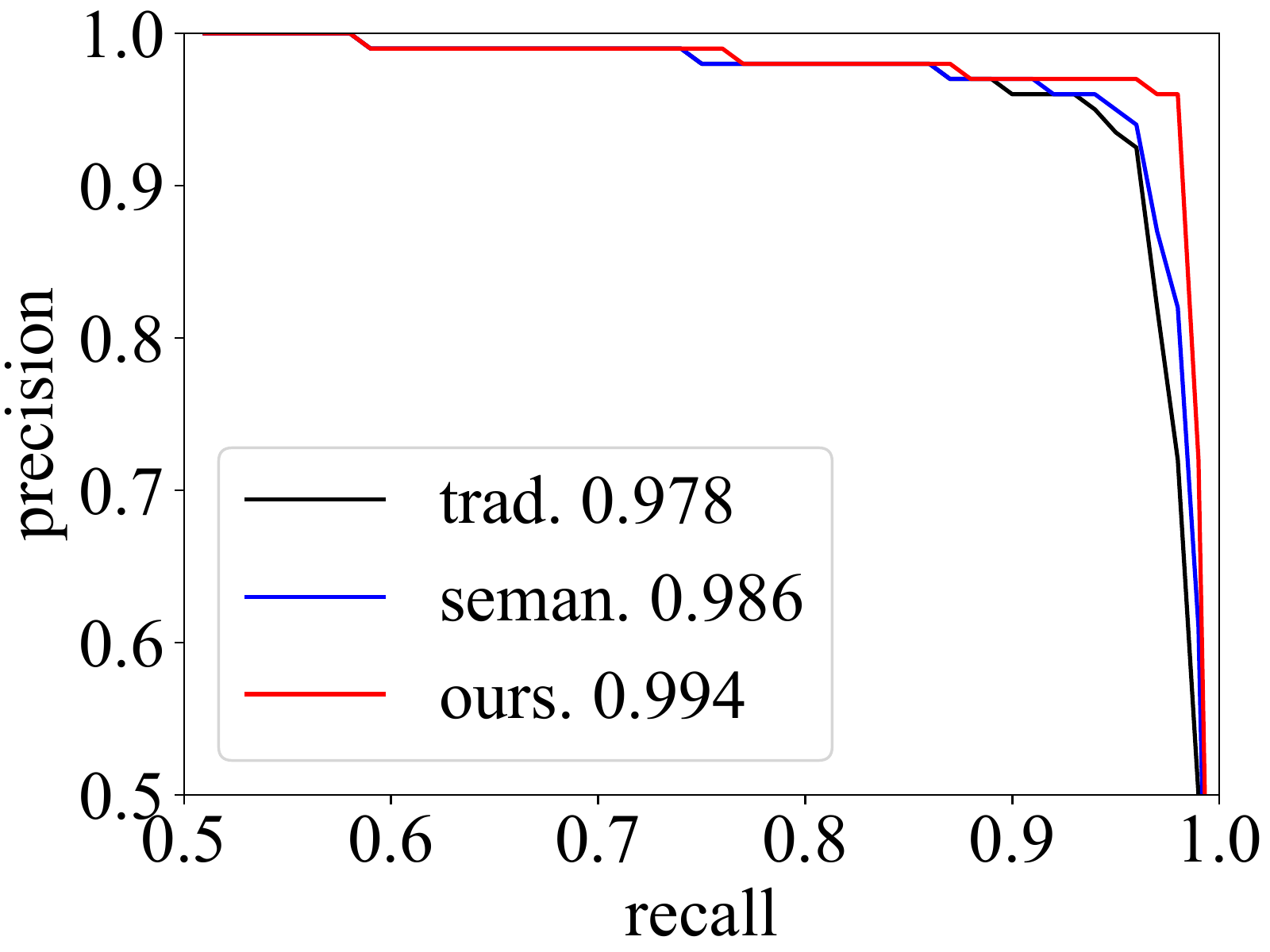}}
	    \vspace{-0.2cm}
	    
	 \subfigure[CityCentre +ORB]{
	    \includegraphics[width=0.48\linewidth]{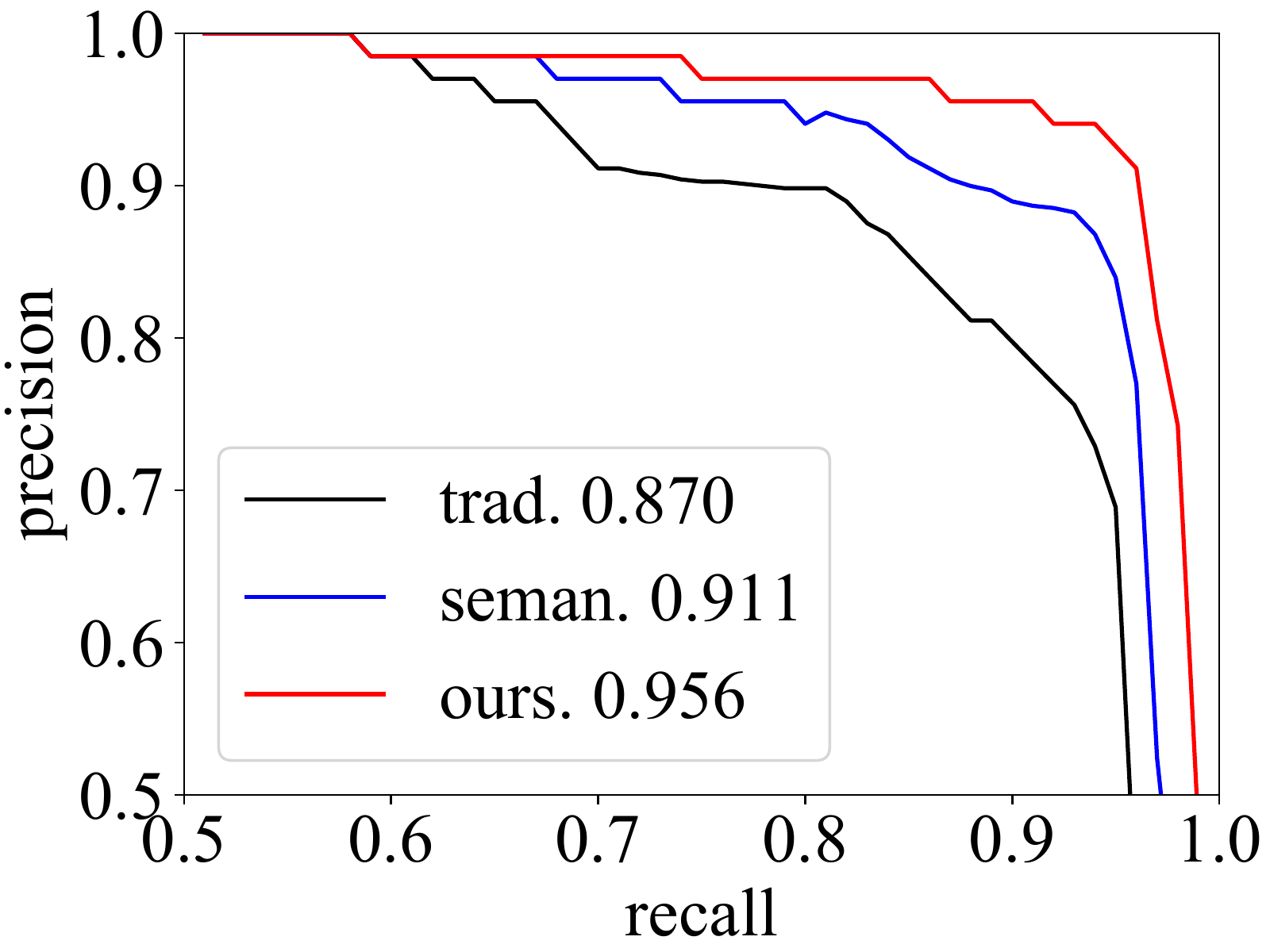}}\subfigure[CityCentre +SuperPoint]{
	    \includegraphics[width=0.48\linewidth]{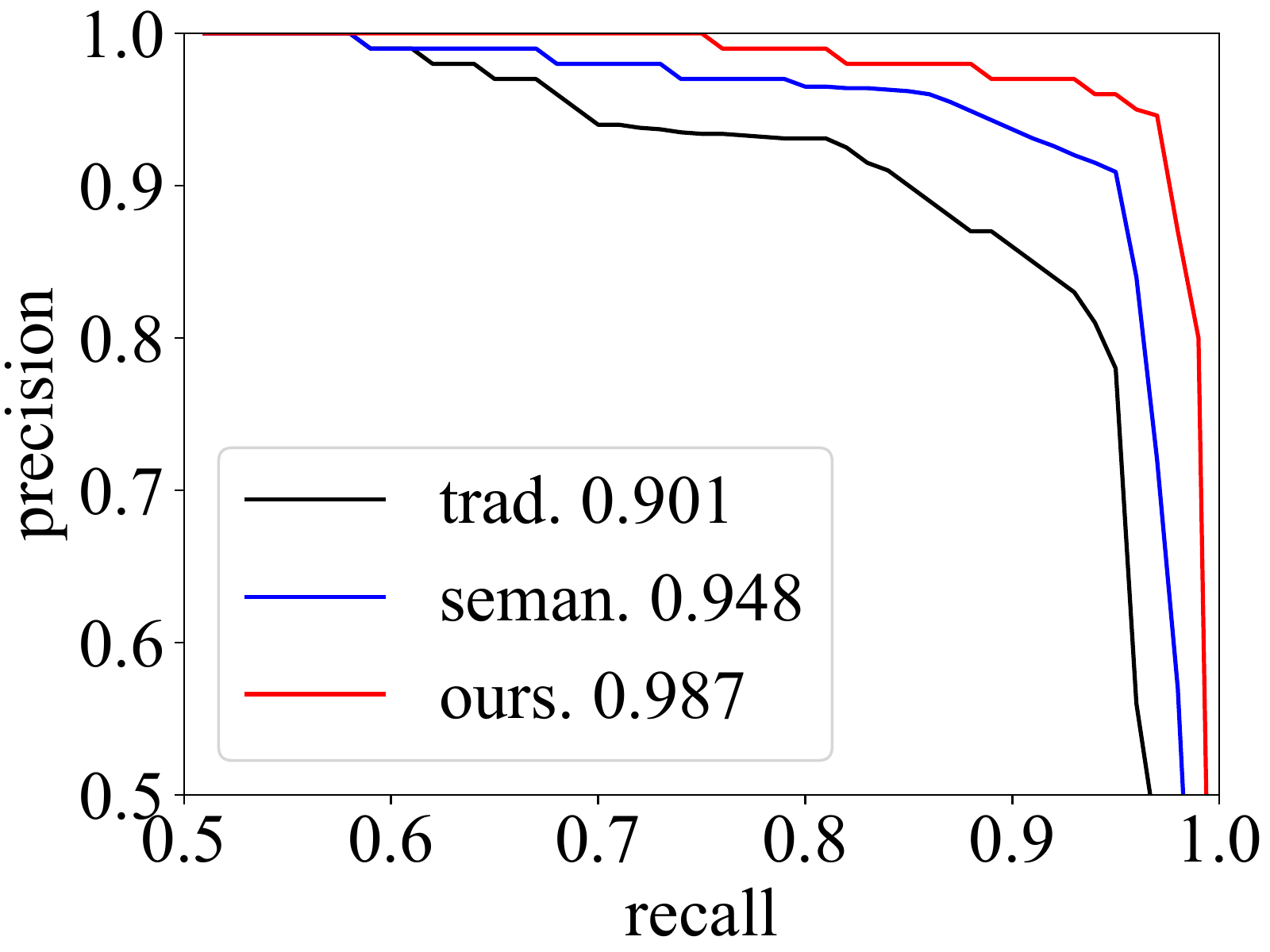}}
	    \vspace{-0.2cm}
\caption{The results in upper row is evaluated on the KITTI-00 dataset using (a) ORB and (b) SuperPoint, while the bottom row is on the CityCentre using (c) ORB and (d) SuperPoint. It should be noticed that the precision and recall axes begin from 0.5.
\label{fig:pr}}
\vspace{-0.5cm}
\end{figure}

In this experiment, we plug our feature selection method into BoW-based place recognition framework \cite{dbow} and then evaluate the effectiveness on KITTI and CityCentre. We test our approach with popularly used sparse and dense local feature, ORB \cite{orb} and SuperPoint \cite{superpoint}, for verifying the versatility and generalization of our method. To comprehensively analyze the effectiveness of our method, we compare with original features and features selected by semantic labels, dubbed as ``trad.'' and ``seman.''.

For original features, we directly use the response value of the characteristic point to select Top-500 features with highest responses. And for the semantic selection methods, we discard features locating on the dynamic objects to avoid the interference from dynamics. Then we also select the Top-500 feature points with the highest response value from remaining features based on the responses of original features. As for our method, the feature points are selected by estimated activation value. 500 features with highest activation are obtained to be further processed. We use RANSAC to estimate the transform matrix between query and loop after place recognition, and we calculate the average re-projection error of inliers as a score to obtain precision-recall curves, as shown in Fig.~\ref{fig:pr}. For quantitative comparison, we also provide area under curve (AUC) in Table~\ref{table:pr}.

\begin{table}[t]
\centering
\caption{Area under curves of compared methods.}
{
\begin{tabular}{c||ccc|ccc}
\toprule
&\multicolumn{3}{c|}{ORB(sparse)}&\multicolumn{3}{c}{SuperPoint(dense)}\\
& trad. & seman. & ours. & trad. & seman. & ours.  \\
\midrule
KITTI00 & 0.989 & 0.992 & \color{red}0.997 & 0.990 & 0.994 & \color{red}0.996\\
KITTI01 & 0.980 & 0.984 & \color{red}0.985 & 0.981 & 0.983 & \color{red}0.985\\
KITTI02 & 0.970 & \color{red}0.977 & 0.976 & 0.967 & \color{red}0.981 & 0.979\\
KITTI05 & 0.965 & 0.971 & \color{red}0.980 & 0.961 & 0.965 & \color{red}0.971\\
KITTI06 & 0.973 & 0.974 & \color{red}0.984 & 0.970 & 0.976 & \color{red}0.988\\
KITTI07 & 0.948 & 0.945 & \color{red}0.950 & 0.951 & 0.949 & \color{red}0.951\\
KITTI08 & 0.963 & 0.962 & \color{red}0.964 & 0.960 & \color{red}0.967 & 0.964\\
KITTI09 & 0.977 & 0.983 & \color{red}0.992 & 0.976 & 0.986 & \color{red}0.994\\
KITTI15 & 0.921 & \color{red}0.930 & 0.928 & 0.935 & 0.939 & \color{red}0.940\\
KITTI16 & 0.935 & 0.947 & \color{red}0.947 & 0.931 & \color{red}0.951 & 0.940\\
KITTI18 & 0.978 & 0.988 & \color{red}0.992 & 0.978 & 0.986 & \color{red}0.990\\
KITTI19 & 0.943 & 0.944 & \color{red}0.952 & 0.946 & 0.948 & \color{red}0.956\\
CityCentre & 0.870 & 0.911 & \color{red}0.956 & 0.901 & 0.948 & \color{red}0.987\\
\bottomrule
\end{tabular}}
\label{table:pr}
\vspace{-0.7cm}
\end{table}

According to the results, we can see that using semantic static/dynamic labels, which is manually defined based on the segmentation results, can sometimes help to discard confusing features so that the performance of place recognition has been improved. However, it sometimes negatively affect the system, such as in KITTI-07 and KITTI-08. It is because that the definition of dynamics and statics is unreasonable and unstable. Moreoever, it cannot focus on distinctive regions based on the texture of images. On the contrary, our proposed DSFeat works well on all the experiments and outperform compared baselines with a lot margins. It concludes that our method has a more stable estimation of dynamic and static attributes and it can detect distinctive regions which helps detection a lot. Additionally, our method can achieve better performance when incorporating with both sparse features and dense features, which is the first feature selection model achieving such goals to the best of our knowledges.

\subsection{SLAM System}
\label{Combine with SLAM system}

\begin{table}[t]
\centering
\caption{The accuracy of SLAM systems.}
{
\begin{tabular}{c||cc|cc}
\toprule
&\multicolumn{2}{c|}{Rotation Error(deg/m)}&\multicolumn{2}{c}{Offset Deviation($\%$)}\\
& ORB-SLAM2 & ours & ORB-SLAM2 & ours  \\
\midrule
KITTI00 & 0.0026 & \color{red}0.0026 & \color{red}0.6943 & 0.7018\\
KITTI01 & 0.0025 & \color{red}0.0020 & 1.6632 & \color{red}1.3842\\
KITTI02 & 0.0029 & \color{red}0.0024 & 0.8394 & \color{red}0.7592\\
KITTI03 & 0.0028 & \color{red}0.0017 & 0.8155 & \color{red}0.7476\\
KITTI04 & 0.0025 & \color{red}0.0015 & 0.5069 & \color{red}0.5051\\
KITTI05 & 0.0017 & \color{red}0.0016 & 0.4188 & \color{red}0.3945\\
KITTI06 & 0.0015 & \color{red}0.0014 & 0.4722 & \color{red}0.4602\\
KITTI07 & \color{red}0.0028 & 0.0029 & \color{red}0.4672 & 0.4742\\
KITTI08 & 0.0031 & \color{red}0.0029 & 1.0438 & \color{red}1.0099\\
KITTI09 & 0.0030 & \color{red}0.0025 & 1.0405 & \color{red}0.8169\\
KITTI10 & 0.0030 & \color{red}0.0028 & 0.6945 & \color{red}0.6419\\
\bottomrule
\end{tabular}
}
\label{table:accuracy}
\vspace{-0.7cm}
\end{table}

For a more comprehensive evaluation, in this experiment, we integrate the proposed method into an practical SLAM system to verify the improvement. We use the ORB-SLAM2 system \cite{orbslam2} as the framework in such experiment and it use ORB as the feature extraction method. Our method is used to select stable features in the feature extraction step of SLAM system. The selected features are then used to track, map, and detect loops, thus affects the performance of overall system. Besides, our provided activation values are also used as a weights in RANSAC procedure during feature matching and pose estimation to improve the accuracy.

\begin{figure}[!t]
\centering{\includegraphics[width=0.47\textwidth]{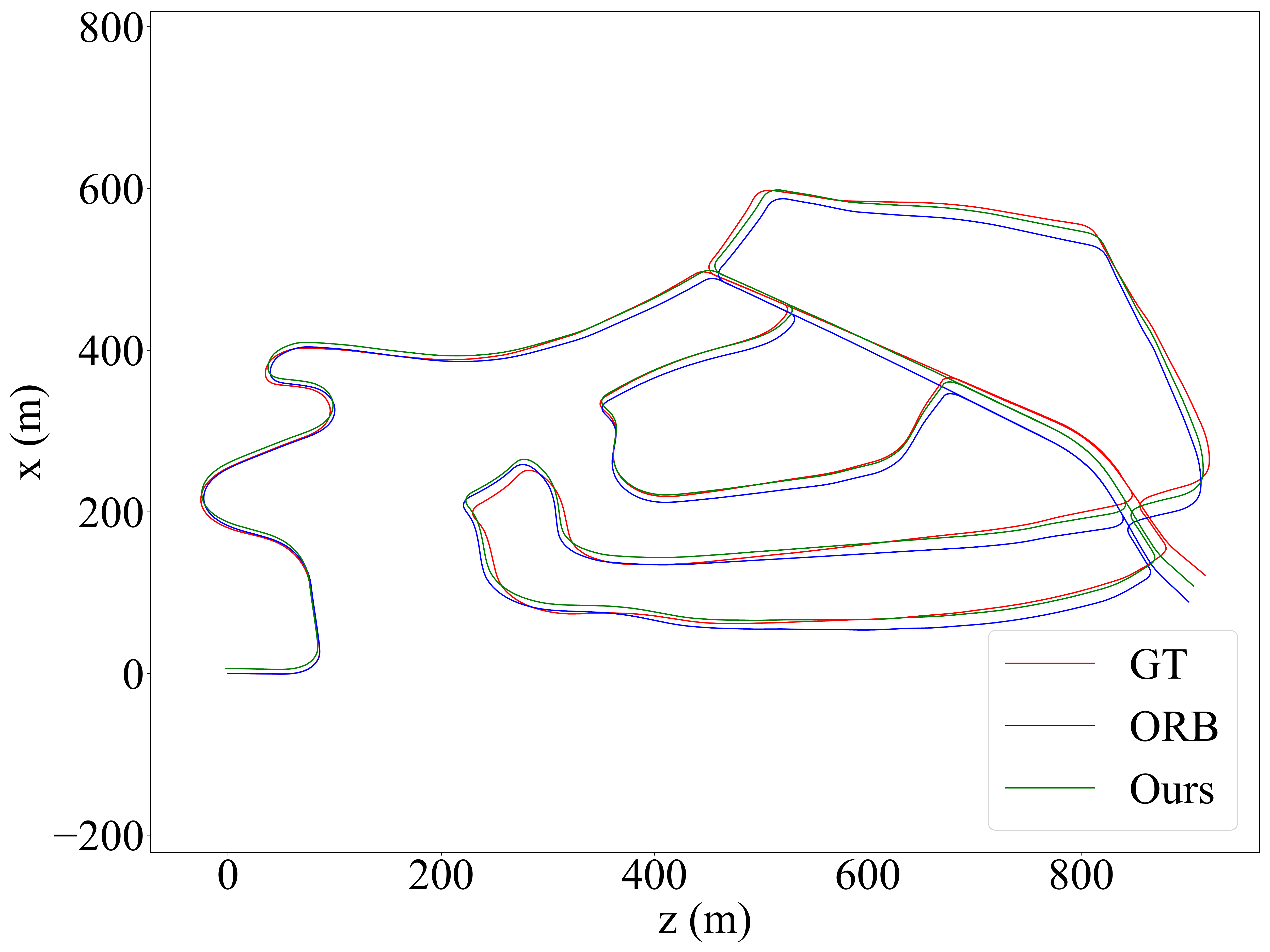}}
\vspace{-0.4cm}
\caption{Visualization of results of our method and ORB-SLAM2 in the KITTI-02 sequence
\label{fig:slam}}
\vspace{-0.2cm}
\end{figure}

We compare the original SLAM system with the one incorporating with our proposed DSFeat. Fig.~\ref{fig:slam} shows an example of trajectories estimated by the slam system and ground-truth in KITTI-02 sequence. Clearly, our estimated trajectory is more accurate than the estimation of original systems which qualitatively concludes the effectiveness of feature selection procedure. Besides, Table~\ref{table:accuracy} records the quantitative results of the average error of rotation and the standard deviation of offset on 11 sequence of KITTI. According to the results, after applying our method to select stable features, the accuracy of the entire SLAM system are improved on almost all the scenes, which concludes the effectiveness and generalization of our proposed approach. 

\begin{figure}[!t]
\centering{\includegraphics[width=0.47\textwidth]{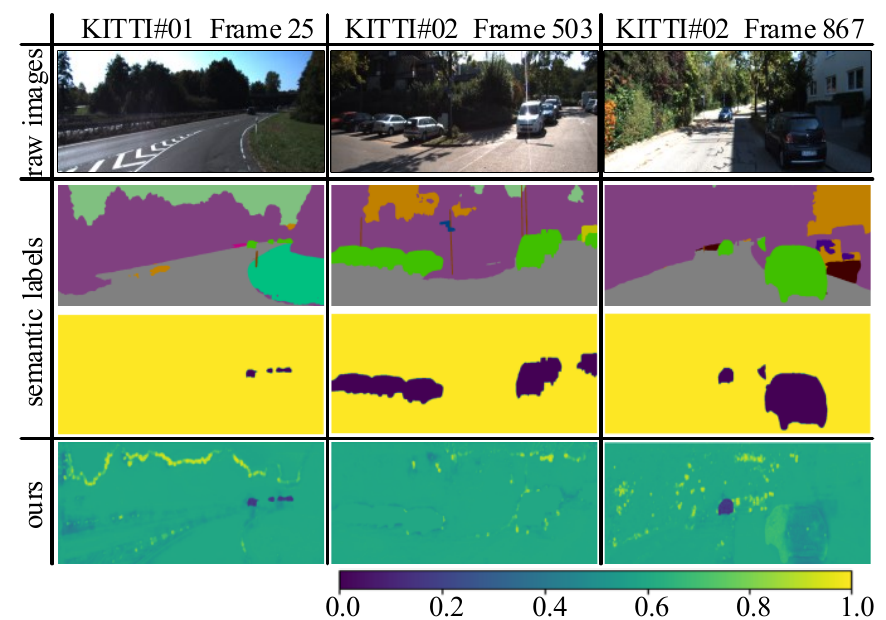}}
\vspace{-0.5cm}
\caption{Qualitative results on KITTI dataset.Pictures on top row are the original images and second ones are the results of semantic segmentation. Third row shows the manually-defined classification of dynamic and static objects based on semantic labels. Activation maps provided by our method are shown in the bottom. Higher activation indicates a static item that should be retained, while regions with lower values are probably dynamics and need to be discarded.}
\label{fig:visual}
\vspace{-0.5cm}
\end{figure}

\subsection{Visualization of Activation Map}
\label{vis}
In order to intuitively visualize the improvement of our proposed DSFeat, we show the semantic labels, semantic static/dynamic map, and our estimated heatmap (activation map) in Fig.~\ref{fig:visual}. Using manually-defined semantic labels to judge the dynamic/static regions is unreasonable and cannot detect distinctive regions. In the figure, all the car is designated to be dynamic according to the semantic segmentation. But cars parked at the roadsides (Fig.~\ref{fig:visual}(b) and Fig.~\ref{fig:visual}(c)) should be static and running ones (Fig.~\ref{fig:visual}(a)) are dynamic. Our method can reliably distinguish them and help to accurately track, map, and detect loops. The selected feature keypoints can be seen in Fig.~\ref{fig:feature}.
\section{Conclusion}
\label{Conclusion}
In this paper, we consider the SLAM problem in complex scenes, and propose a method of using activation map for feature point selection. First, we improve the backbone of U-net and propose a network structure for generating one-dimensional activation maps. Second, we specifically design corresponding dis-similarity measurement function for dense and sparse features. Third, we combine the strong supervision from semantic information with the self supervision from matching information as a hybrid loss function to train the model, and regulate the ratio between various supervision based on the training process. Finally, we use the estimated activation map to select stable feature and add weights in the RANSAC algorithm. The experimental results show that our approach effectively improves the effectiveness of image retrieval in complex scenes, and also significantly improves the accuracy of localization in the SLAM system. And It should be noticed that the proposed feature selection can work well with both sparse features and dense features, and its generalization is concluded in various scenes. The proposed DSFeat can be readily pluggable into any odometry or SLAM inplements based on local features and we will release the related codes later for further studies.

\bibliographystyle{IEEEtran} % We choose the "plain" reference style
\bibliography{refs} % Entries are in the "refs.bib" file
\end{CJK*}
\end{document}